\documentclass[10pt,conference,a4paper]{IEEEtran}
\usepackage{cite}

\ifCLASSINFOpdf
  \usepackage[pdftex]{graphicx}
  \graphicspath{{./Images/}}
\else
\fi
\usepackage{xcolor}
\usepackage{amsmath,amssymb} 
\usepackage{multicol}
\usepackage{multirow}
\usepackage{subcaption}
\usepackage{siunitx}
\usepackage{graphicx}
\begin{document}
%

\title{Compact CNN Structure Learning \\ by Knowledge Distillation}



%
\author{
\IEEEauthorblockN{Waqar Ahmed\IEEEauthorrefmark{1}\IEEEauthorrefmark{2},
Andrea Zunino\IEEEauthorrefmark{3},
Pietro Morerio\IEEEauthorrefmark{1} and
Vittorio Murino\IEEEauthorrefmark{1}\IEEEauthorrefmark{3}\IEEEauthorrefmark{4}
}
\IEEEauthorblockN{\{waqar.ahmed, pietro.morerio, vittorio.murino\}@iit.it,
andrea.zunino@huawei.com
}
\IEEEauthorblockA{\IEEEauthorrefmark{1}Pattern Analysis \& Computer Vision (PAVIS), Istituto Italiano di Tecnologia, Genova, Italy}
\IEEEauthorblockA{\IEEEauthorrefmark{2}Dipartimento di Ingegneria Navale, Elettrica, Elettronica e delle Telecomunicazioni, University of Genova, Italy}
\IEEEauthorblockA{\IEEEauthorrefmark{3}Ireland Research Center, Huawei Technologies Co. Ltd., Dublin, Ireland}
\IEEEauthorblockA{\IEEEauthorrefmark{4}Dipartimento di Informatica, University of Verona, Italy}
}


\maketitle

\begin{abstract}
The concept of compressing deep Convolutional Neural Networks (CNNs) is essential to use limited computation, power, and memory resources on embedded devices. However, existing methods achieve this objective at the cost of a drop in inference accuracy in computer vision tasks.
To address such a drawback, we propose a framework that leverages knowledge distillation along with customizable block-wise optimization to learn a lightweight CNN structure while preserving better control over the compression-performance tradeoff. Considering specific resource constraints, e.g., floating-point operations per inference (FLOPs) or model-parameters, our method results in a state of the art network compression while being capable of achieving better inference accuracy.
In a comprehensive evaluation, we demonstrate that our method is effective, robust, and consistent with results over a variety of network architectures and datasets, at negligible training overhead. In particular, for the already compact network MobileNet\_v2, our method offers up to 2$\times$ and 5.2$\times$ better model compression in terms of FLOPs and model-parameters, respectively, while getting 1.05\% better model performance than the baseline network.
\end{abstract}


%
\IEEEpeerreviewmaketitle

\section{Introduction}
Recent years have seen remarkable performance breakthroughs achieved in machine learning \cite{ahmed2020performance} and computer vision applications using deep Convolutional Neural Networks (CNNs) \cite{he2016deep,pmlr-v80-kalchbrenner18a,Ronneberger2015UNetCN}. 
However, the CNNs are computationally expensive, memory-intensive, and power hungry. Therefore, extraordinary inference speed, throughput, and energy efficiency are required to meet the real-time application's demands running on resource-constrained devices such as drones, robots, smartphones, and wearable devices \cite{he2015convolutional}. 

Researchers have demonstrated the possibility of using an automated architecture search approach to discover an optimal CNN structure for the task of interest.  Yet, it is an impractical method that requires a huge architecture searching time in finding a reasonable solution due to the combinatorially large search space \cite{zoph2016neural,pham2018efficient}.
Another possible proposed direction is to design lightweight CNN architectures that typically requires expensive, frequently manual, trial-and-error exploration to find a good solution. However, CNN customized for a particular task fails to maintain the required performance in other tasks: thus, a similar exercise is needed to target every new problem. Nevertheless, former methods do not consider precise resource constraints (\textit{i.e.} FLOPs and model-parameters) and are not scalable for growing task complexity.

To overcome these challenges, we propose a powerful and adaptive method for learning an optimal network structure for the task of interest. Our approach advances the spirit of recently proposed method \textit{MorphNet} \cite{gordon2018morphnet} which has the advantage of being fast, scalable and adaptive to specific resource constraints (e.g., FLOPs or model-parameters). Most importantly, it \textit{learns} network structure during training. However, the current optimization technique has an intrinsically \textit{biased concentration} that either pushes the optimizer to focus on high-resolution layers (towards network input) or focus more on low-resolution layers (towards network output) when optimized for FLOPs or model-parameters, respectively. 
Consequently, it leads to an sub-optimal network structure and reduced model performance. Thus, \cite{gordon2018morphnet} employs a width multiplier to uniformly expand all layers to improve model performance which eventually results in a similar resource-intensive network structure.

\begin{figure*}[ht!]
\centering
\includegraphics[width=0.9\linewidth]{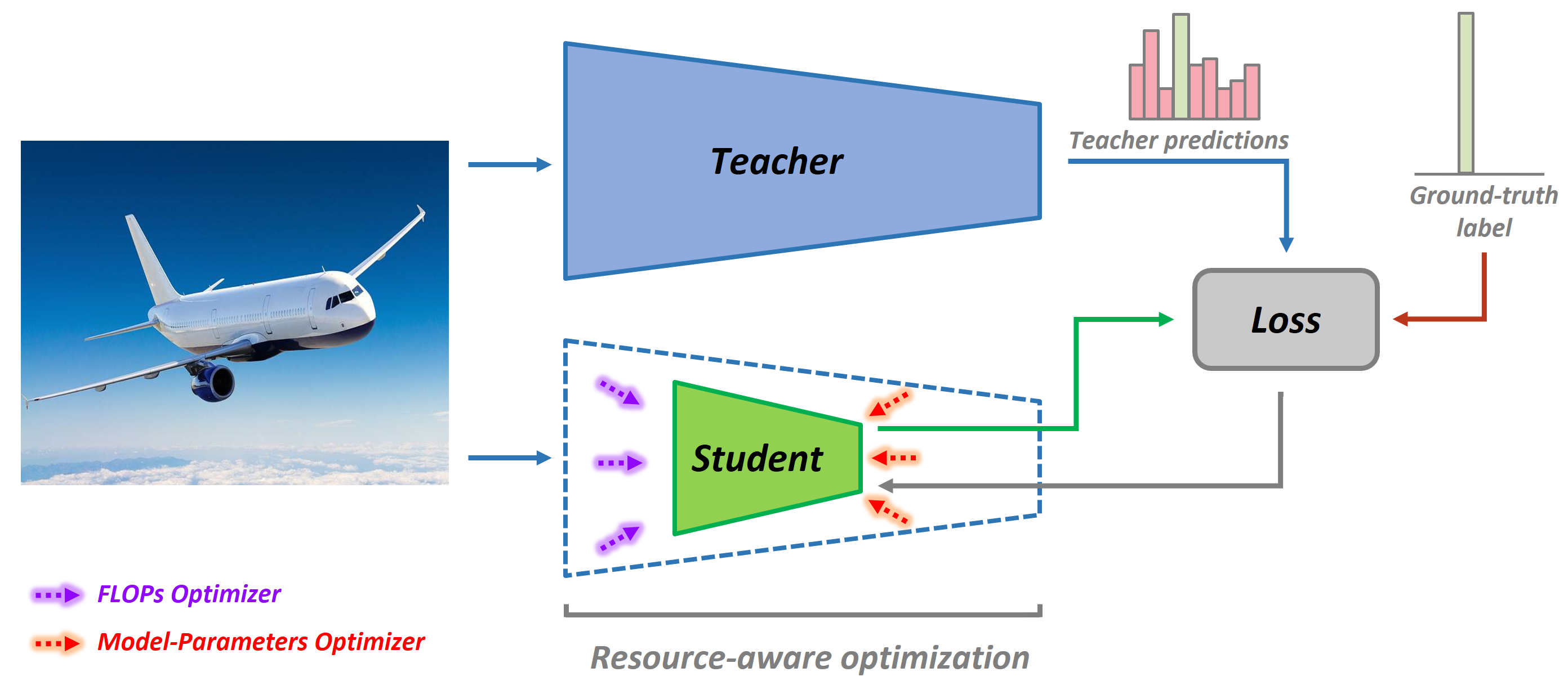}
\caption{\textbf{Overview of the proposed method.} Our method assumes a teacher network which is usually the network to be compressed itself. The resource-aware optimization employs FLOPs and model-parameters optimizers on suitable parts of the student network with respective budget constraints. While it relaxes the task complexity, privileged information imposes control over predictions to preserve superior model performance during network structure learning.
}
\label{fig:pipeline}
\end{figure*}

To mitigate these shortcomings, our method employs \emph{Resource-aware Optimization} augmented with the \emph{Privileged Information (PI)} technique in a student-teacher scheme (see Figure~\ref{fig:pipeline}). 
The proposed resource-aware optimization breaks down the seed network in smaller instances which curtails task complexity to learn better end-to-end network structure. Eventually, it enables customized optimization of each stage of the network with specific budget constraints.
As several works have already proved PI's potential in improving model performance \cite{vapnik2015learning,sharmanska2013learning,LopSchBotVap16}, in our case, it augments our method's capability by imposing control over model performance during optimization considering the teacher network performance as a target. This facilitates the optimizer to maintain high model performance while learning the lightweight network structure. Note that our method does not apply network expansion at all, and the student network to be compressed utilizes PI extracted almost for free from the uncompressed network itself.

The hybrid of the above-mentioned strategies leads to a superior and consistent compression results over a variety of network architectures (e.g., ResNet101 and MobileNet\_v2) and datasets (e.g., CIFAR-10, CIFAR-100, and ImageNet).
The proposed method is novel, effective and carefully devised to target specific limitations of the existing method. As a result, an optimal network structure is discovered which is also capable of delivering better model performance.
In particular, for image recognition task using the already compact network MobileNet\_v2 on CIFAR-10 benchmark, our method achieves 2$\times$ and 5.2$\times$ better model compression than \cite{gordon2018morphnet}, in terms of FLOPs and model-parameters, respectively (see Figure~\ref{fig_CIFAR10}). 
Especially, the resultant network delivers 80.38\% classification accuracy which is 1.05\% better than the baseline teacher network.

The rest of the paper is organized as follows. Section~\ref{rel_work} reviews related works. Section~\ref{method} introduces the background knowledge, necessary notations, and our proposed approach. Section~\ref{exps} discusses our experimental setup and Section~\ref{result_sec} presents the results. Finally, Section~\ref{conclus} sketches the conclusion.

\section{Related Work}\label{rel_work}
Design of computation, memory, and power-efficient CNNs are needed to deploy deep learning applications on embedded devices such as drones, robots, and smartphones \cite{he2015convolutional}. One way of achieving this objective is to transfer knowledge from a deep, wide and complex \emph{teacher} network to a shallow \emph{student} network \cite{hinton2015distilling,LopSchBotVap16,sun2019patient,wang2019private}. 
In principle, a teacher network is trained in advance. Afterward, a lightweight student network is trained to mimic the behavior of the teacher network in an equally effective manner. Also, a quantized distillation approach was proposed in \cite{polino2018model}, whereas \cite{belagiannis2018adversarial} suggests an adversarial learning process for model compression. However, in existing works, distilling the generalization ability of complex teacher into a smaller network cost superfluous FLOPs and model-parameters depending on \emph{predefined fixed structure} of the student. Differently, we propose an approach leveraging privileged information to dynamically learn an optimal network structure of a student while respecting the given resource constraints for the task of interest.

Some recent works achieve model compression by pruning redundant connections \cite{carreira2018learning,molchanov2019importance} or using low-precision/quantized weights \cite{polino2018model,wu2016quantized}. While others propose a precise design of efficient CNN architectures by inverting residual connections between the thin bottleneck layers \cite{sandler2018mobilenetv2}, grouping point-wise and depth-wise dilated separable convolutions \cite{mehta2019espnetv2}, or
utilizing pointwise group convolution and channel shuffle \cite{zhang2018shufflenet}. However, the designing of efficient CNNs approaches is not scalable and requires extensive human efforts to target every new problem, dataset or platform.
On the contrary, our proposed method automatically \emph{learns} a lightweight network (student) structure, sufficient to deliver a comparable performance of a given large and complex (teacher) network.

Some related methods exist, such as \cite{Yang_2018_ECCV} which progressively simplifies a pre-trained CNN by generating network proposals during training until the resource budget is met. AutoGrow \cite{wen2019autogrow} automates the process of depth discovery in CNNs by adding new layers in shallow seed networks until the required accuracy is observed. 
However, our proposed approach is inspired by MorphNet \cite{gordon2018morphnet}, an open-source tool for learning network structure based using resource weighted sparsifying regularizer. Among all state-of-the-art works, this method is fast, scalable and adaptable to specific resource constraints (e.g., FLOPs or model-parameters). 
However, the optimization comes with a drawback, \textit{i.e.} the more you iterate, the more you observe a drop in accuracy. To recover performance loss, all layers are uniformly expanded using a width multiplier which may lead to an improved but large resource-consuming network structure. 
To mitigate these limitations, our method utilizes \emph{Privileged Information} along with \emph{Resource-aware Optimization} to improve the network structure learning process. Without any network expansion, the proposed method offers superior FLOPs and model-parameters reduction along with better model performance.

\section{Method}\label{method}
We propose a framework that learns an optimal CNN structure to efficiently target the task of interest, considering allowed resource constraints e.g., FLOPs and model-parameters while preserving high model performance. 
We advance the spirit of MorphNet which is based on a training procedure to optimize CNN structure. Since it does not represent a simple pruning or post-processing technique, the method is well suited for our task. 
The model compression method of MorphNet \cite{gordon2018morphnet} relies on a regularizer $\boldsymbol{\mathcal{R}}$. It induces sparsity in activations by putting greater cost $\boldsymbol{\mathcal{C}}$ on neurons contributing to either FLOPs or the model-parameters. 
The network sparsity is measured on the basis of batch normalization scaling factor $\gamma$ associated with each neuron \textit{i.e,} if $\gamma$ lies below than the user-defined threshold, the corresponding neuron is considered as dead and can be discarded (since its scale is negligible). 
Both the FLOPs and model-parameters are influenced by the particular layer associated with matrix multiplications - \textit{i.e.} convolutions.
This makes sense, as the lower layers of the neural network are applied to a high-resolution image, and thus consume a large number of the total FLOPs. Whereas, the upper layers typically comprises of larger number of channels and thus contain abundant weight matrices. We can define separate cost functions as follows:
\begin{equation}\label{eq:flops_cost}
    \boldsymbol{\mathcal{C}}_{FLOP} = \sum_{k=1}^{K}[C^k_{in} * (w^k)^2 * C^k_{out} * S^k_{out}]
\end{equation}
\begin{equation}\label{eq:size_cost}
\boldsymbol{\mathcal{C}}_{PARAM} = \sum_{k=1}^{K}[C^k_{in} * (w^k)^2 * C^k_{out}]
\end{equation}
where $K$ is total number of layers and $k$ is the layer index, $w^2$ is the kernel size, $C_{in}$ is the number of input channels, $C_{out}$ is the number of output channels, and  $S_{out}$ is the size of the output layer (\textit{i.e.} the number of times the kernel is applied). In this work, we propose to use two different regularizers $\boldsymbol{\mathcal{R}}_{FLOP}$ and $\boldsymbol{\mathcal{R}}_{PARAM}$, depending on the resources being optimized in a particular stage of the network. So the optimization problem is equivalent to applying a penalty on the loss as follows:
\begin{equation}\label{eq:morphnet_optimization}
\displaystyle\min_{\substack{\theta}} \boldsymbol{\mathcal{L}}(\theta) + \alpha\boldsymbol{\mathcal{C}}_j(\theta) ,  \;  j = \{ FLOP, PARAM  \}.
\end{equation}
where $\boldsymbol{\mathcal{C}}$ is a function of the model-parameters $\theta$ and the hyperparameter $\alpha$ regulates the resource optimization intensity. For comparison, we refer to MNF and MNP as the original MorphNet method which optimizes the entire network for FLOPs and model-parameters, respectively. 

\begin{figure}[!t]
\centering
\includegraphics[width=1\linewidth, height=6.5cm]{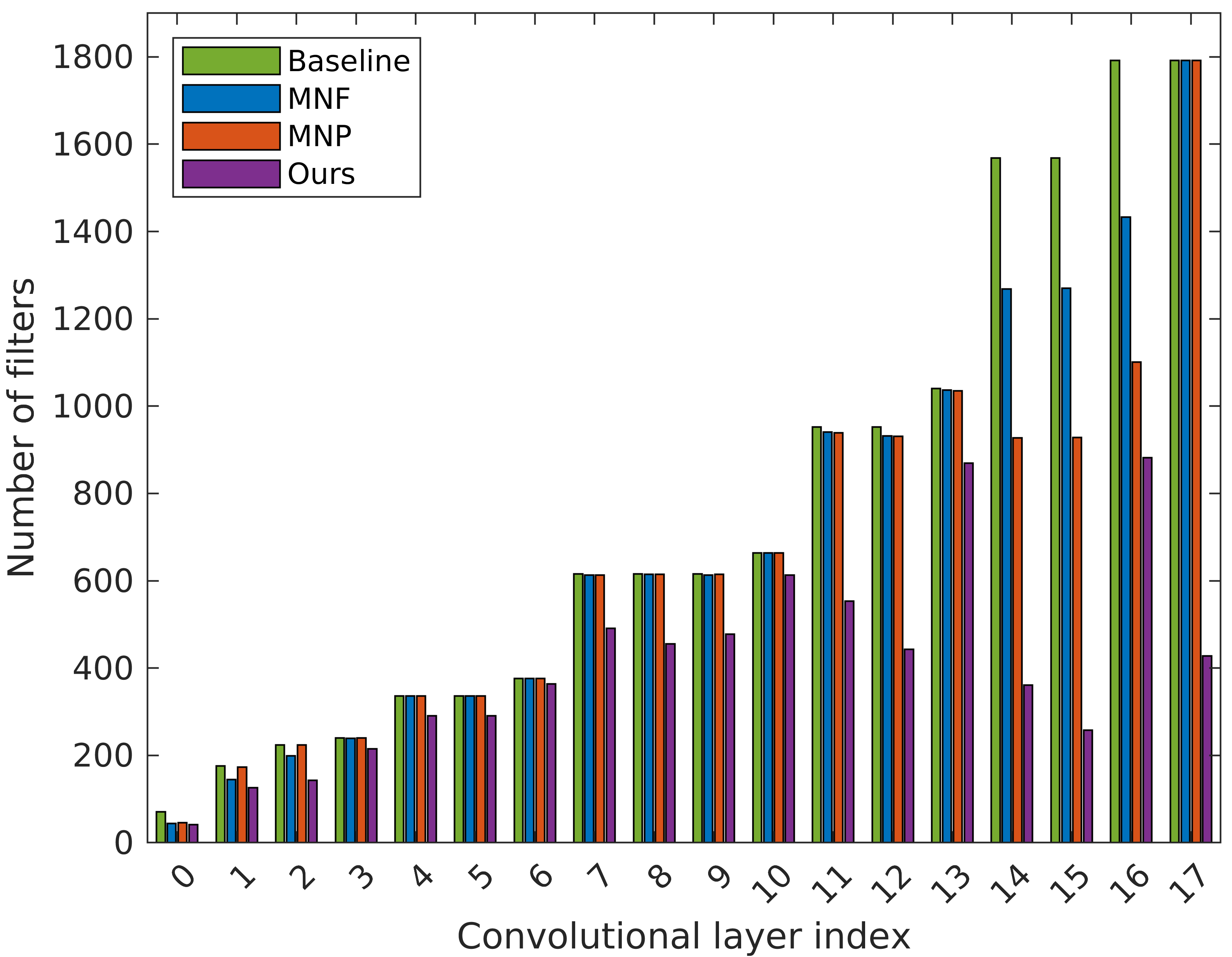}
\caption{\textbf{MobileNet\_v2 structure optimized on CIFAR-100.} The percentage of dropped filters from each convolution stage of MobileNet\_v2 is presented. Comparing with baseline (outright number of filters), it can be observed that FLOP regularizer (MNF) of the existing method tends to remove filters from the lower layers near the input, whereas the model-parameters regularizer (MNP) tends to remove more filters from upper layers near the output. On the contrary, in terms of model compression, our proposed method clearly outperforms the existing method by a large margin over all stages of the network. Please refer to Figure~\ref{fig_CIFAR100} for the accuracy comparison.}
\label{fig_arch}
\end{figure}

The network structure obtained using stand-alone MorphNet costs a significant drop in model performance. This happens because its optimization is either based on Eq.~\eqref{eq:flops_cost} or \eqref{eq:size_cost} that leads to a biased concentration that either forces the optimizer to focus on high-resolution layers (towards network input) or focus on low-resolution layers (towards network output) when optimized for FLOPs or model-parameters, respectively (see Figure~\ref{fig_arch}). Consequently, learning structure of the entire complex network with such a biased method leads to a sub-optimal solution and a reduced model performance.
To recover performance loss, the existing method uniformly expands all layer sizes using a width multiplier which may lead to a better but large resource-consuming network structure.

To overcome these challenges, our method employs \emph{Resource-aware Optimization} augmented with the \emph{Privileged Information (PI)} technique in a student-teacher scheme (see Figure~\ref{fig:pipeline}). 
The resource-aware optimization breaks the complex task of learning the entire CNN structure into comparatively simpler sub-tasks. Accordingly, with the reduced complexity (to find the structure of sub-network only), it enables customized optimization of each stage of the network with specific budget constraints. Eventually, our approach discovers a global network structure, lighter than the original end-to-end solution. 

In addition to that, the privileged information framework \cite{LopSchBotVap16}, augments our method's capability by imposing control over model performance during optimization considering the teacher network performance as a target. This facilitates the optimizer to maintain high model performance while learning the optimal network structure. Note that our method does not apply network expansion at all, and the student network to be compressed utilizes PI extracted almost for free from the uncompressed network itself. In this way, the impact on performance is also accounted along with the existing sparsity measure that helps the optimizer to remove only the least significant neurons from the network. 

\subsection{Leveraging Privileged Information}
A pre-trained teacher network $f_{t}$ is cloned to serve as a student network $f_{s}$ (a network to be compressed). 
Although the architecture of the teacher can be different, the choice of the same architecture eliminates the requirement of training an additional teacher network. 
As depicted in Figure~\ref{fig:pipeline}, the structure of the student network is optimized to meet the required resource budget while taking advantage of the soft predictions (from the teacher) along with the ground-truth labels. This forces the student network to keep mimicking baseline predictions during optimization. 
In principle, both networks are trained to achieve the same task - \textit{i.e.} infer the identical class of input image $x^i$. The training is accomplished by minimizing the following cross-entropy loss:
\begin{equation}\label{eq:soft_pred}
\boldsymbol{\mathcal{L}}(\boldsymbol{\theta})=\frac{1}{N}\sum_{i=1}^{N}l\big(y^{i},f(\mathbf{x}^{i},\boldsymbol{\theta})\big)
\end{equation}
where $\boldsymbol{\theta}$ are parameters of the model, $y^{i}$ is the ground-truth label of sample $i$, $f(\cdot)$ symbolizes the activation function, $\mathbf{x}^{i}=\{x_{1}^{i},x_{2}^{i},\dots,x_{m}^{i}\}\in\mathbb{R}^{m}$ denotes a training sample, and $l$ is a loss to measure the prediction error.
During the inference of $y^i$ given $x^i$, $i=1,\dots,N$, the student network leverages privileged information $z^i$ about the sample $(x^i,y^i)$. Such additional information is derived from the teacher model's prediction:
\begin{equation}\label{eq:PI}
z^i = \sigma(f_t(x^i)/T)
\end{equation}
where $\sigma$ symbolizes the softmax operator and $f_t(x^i)$ refers to the teacher logits. Thus, the student network is trained according to the following optimization problem:
\begin{equation}\label{eq:distillation}
\begin{split}
f_s = \displaystyle\arg \min_{\substack{f \in F_s}} \frac{1}{N} \sum_{i=1}^{N}[(1-\lambda)l(y^i,\sigma(f(x^i))) \\
+  \lambda l(z^i,\sigma(f_t(x^i)/T))].
\end{split}
\end{equation}
The parameter $T$ regulates the amount of smoothness applied to logits. This not only reveals commonalities and differences between classes to be discriminated but also exploits the true potential of the soft labels \cite{LopSchBotVap16}. ${F_s}$ is the student function hypothesis space and the parameter ${\lambda \in [0, 1]}$ is the imitation factor, controlling the student to mimic the teacher vs. to predict the ground-truth label. 
Therefore, the proposed approach is modeled by incorporating Eq.~\eqref{eq:distillation} into MorphNet's minimization equation (in Eq.~\eqref{eq:morphnet_optimization}). Thus we get the following optimization problem:
\begin{equation}\label{eq:PI_and_morphnet}
\begin{split} 
\min_{\theta} \frac{1}{N} \sum_{i=1}^{N}[ (1-\lambda)  \, & l(y^i,\sigma(f(x^i, \theta)/T)) \\
+  \lambda  \, & l(z^i,\sigma(f_t(x^i, \theta)/T)) \\
+   \alpha  \, & \boldsymbol{\mathcal{C}}_j(\theta)], 
\quad  j = \{ FLOP, PARAM  \},
\end{split}
\end{equation}
where the standard cross-entropy loss incorporates both, the privileged information and the MorphNet optimizations. In principle, incorporating privileged information from the uncompressed method is sufficiently general to be applied to any compression algorithm which can be expressed in the form of Eq.  \eqref{eq:morphnet_optimization}.

\subsection{Resource-aware optimization}\label{block}
In CNNs, the input image is processed by progressively reducing the feature map's resolution while increasing the number of filters to be applied along with the network. Therefore, lower layers carry higher \textit{FLOPs Cost}, while higher layers account for huge model-parameters (see Eq.~\eqref{eq:flops_cost}\&\eqref{eq:size_cost}). 

However, Figure~\ref{fig_arch} shows that layers of MobileNet\_v2 are reduced differently according to the type of optimization (MNF or MNP) performed during training. This suggests that different optimizers on lower and upper layers are needed in order to be more effective in network structure learning. For this reason, we propose a resource-aware optimization scheme to optimize different layers of the network using suitable optimizer which leads to following modification to Eq. \eqref{eq:PI_and_morphnet}:
\begin{equation}\label{eq:blockwise}
\begin{split} 
\min_{\theta_1}  \min_{\theta_2} \frac{1}{N} \sum_{i=1}^{N}[ (1-\lambda)  \, & l(y^i,\sigma(f(x^i, \theta_1, \theta_2)/T)) \\
+  \lambda  \, & l(z^i,\sigma(f_t(x^i, \theta_1, \theta_2)/T)) \\
+   \alpha  \, & (\boldsymbol{\mathcal{C}}_{FLOP}(\theta_1) +  \boldsymbol{\mathcal{C}}_{PARAM}(\theta_2) ) ],  
\end{split}
\end{equation}
where $\theta_1 \cup \theta_2 = \theta, \; \theta_1 \cap \theta_2 = \varnothing $ is a partition of the weights parametrizing the lower and upper layers of the network, respectively. Specifically, we propose a configuration in which the first half of the network is optimized for FLOPs and the second half is optimized for model-parameters. 
\begin{figure*}
\captionsetup{font=normalsize}
\begin{subfigure}{\textwidth}
\captionsetup{font=normalsize}
\centering
\includegraphics[width=0.45\linewidth,height=5.0cm]{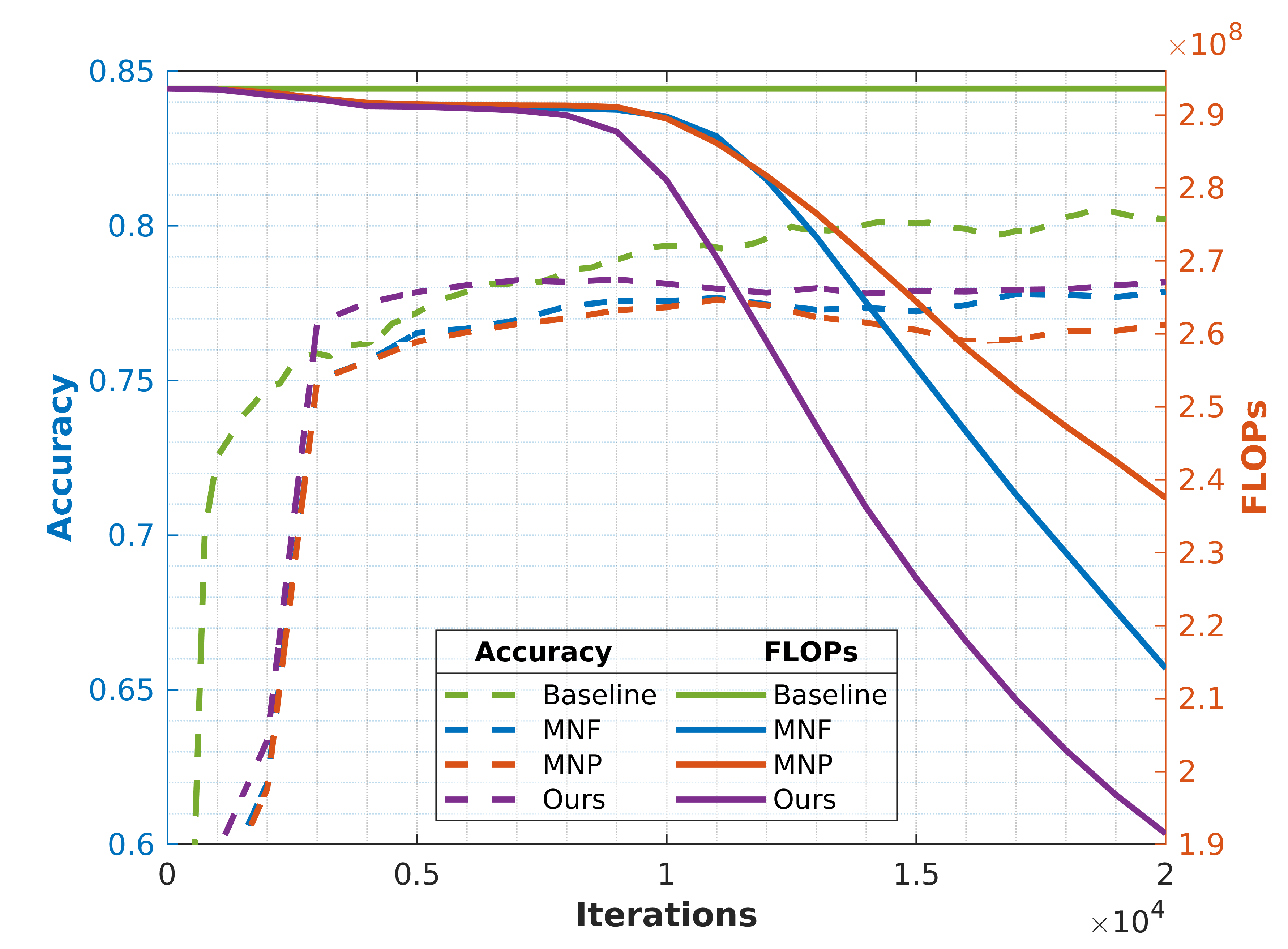}
\includegraphics[width=0.45\linewidth,height=5.0cm]{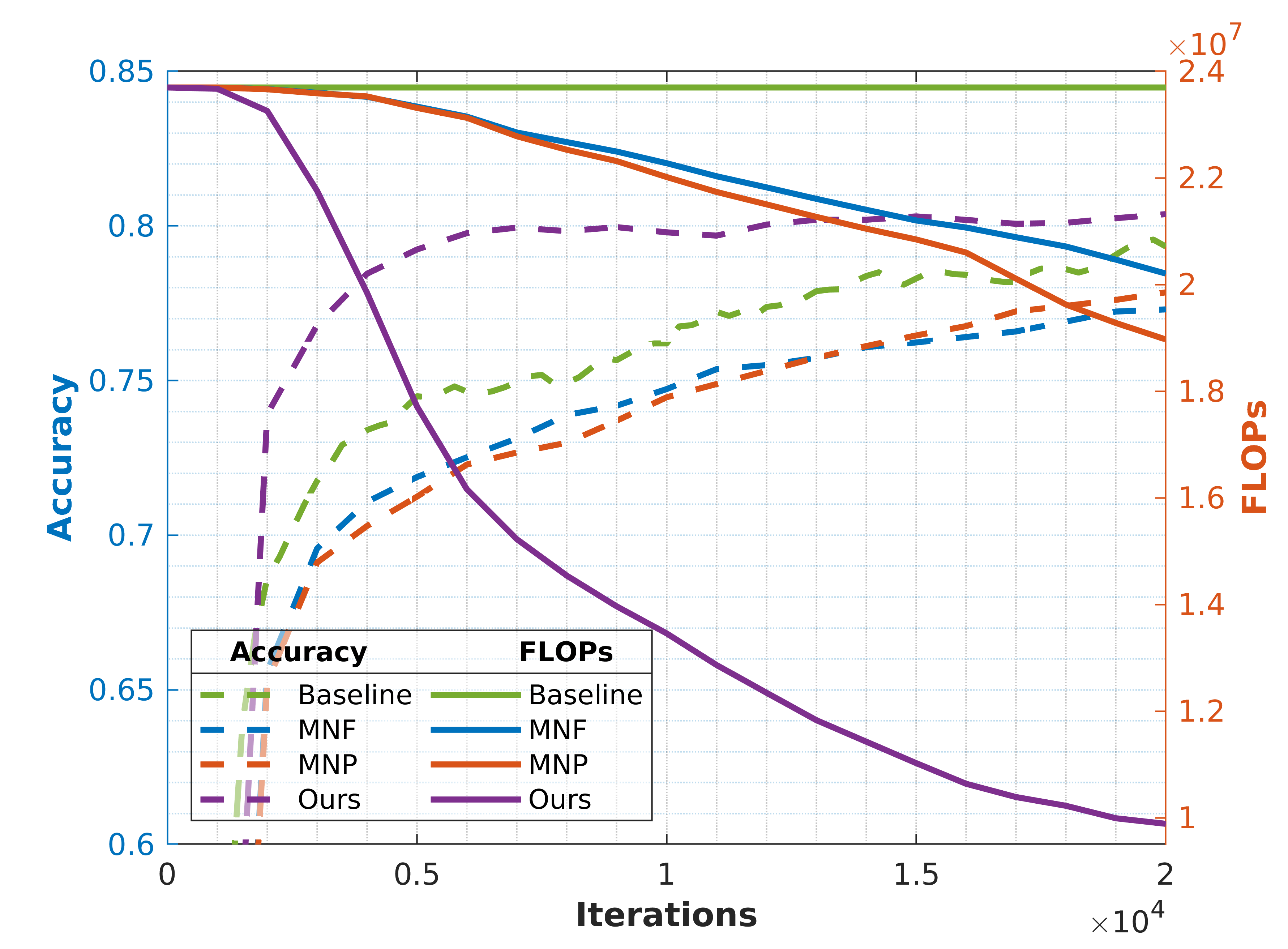}

\includegraphics[width=0.45\linewidth,height=5.0cm]{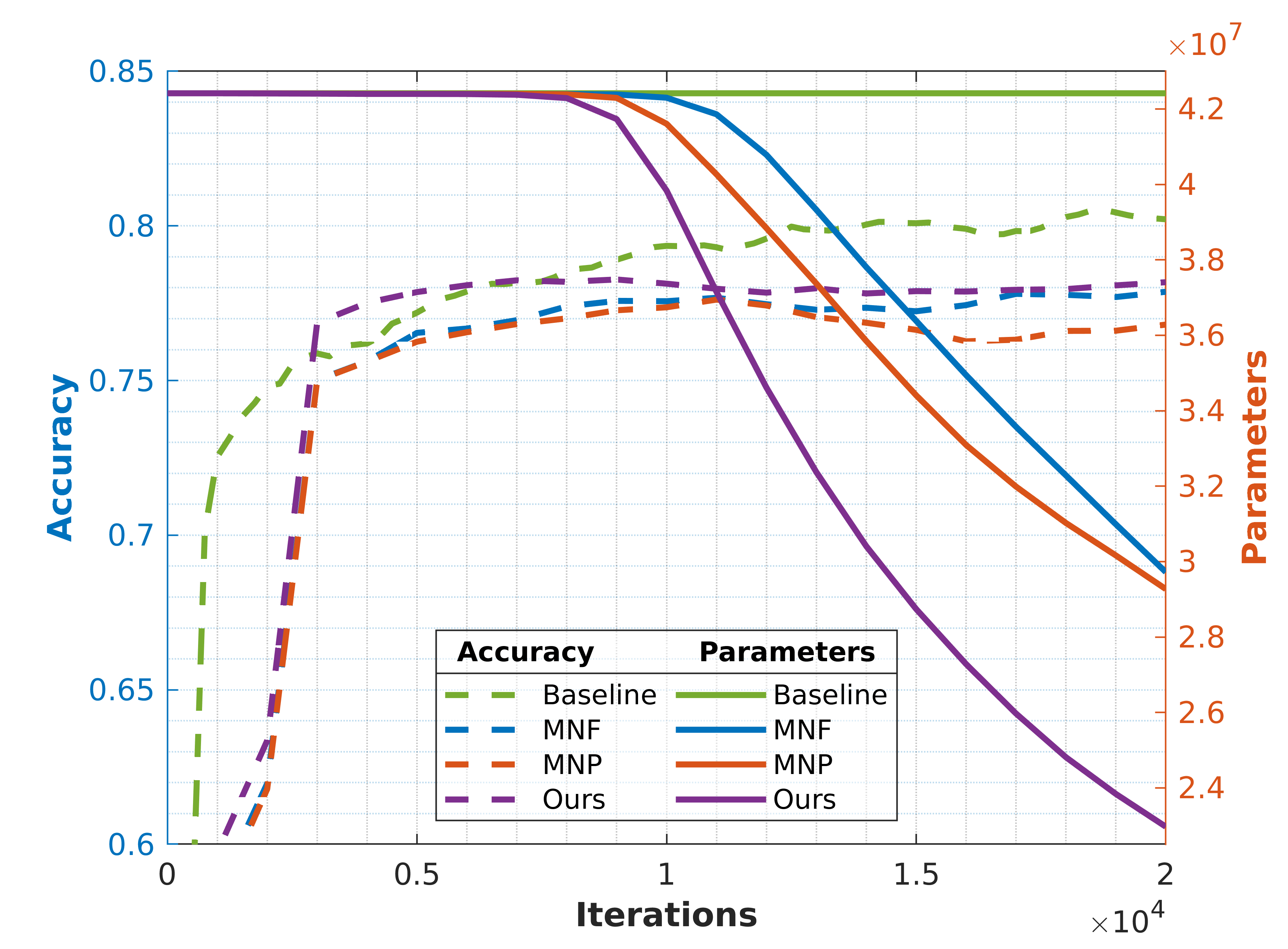}
\includegraphics[width=0.45\linewidth,height=5.0cm]{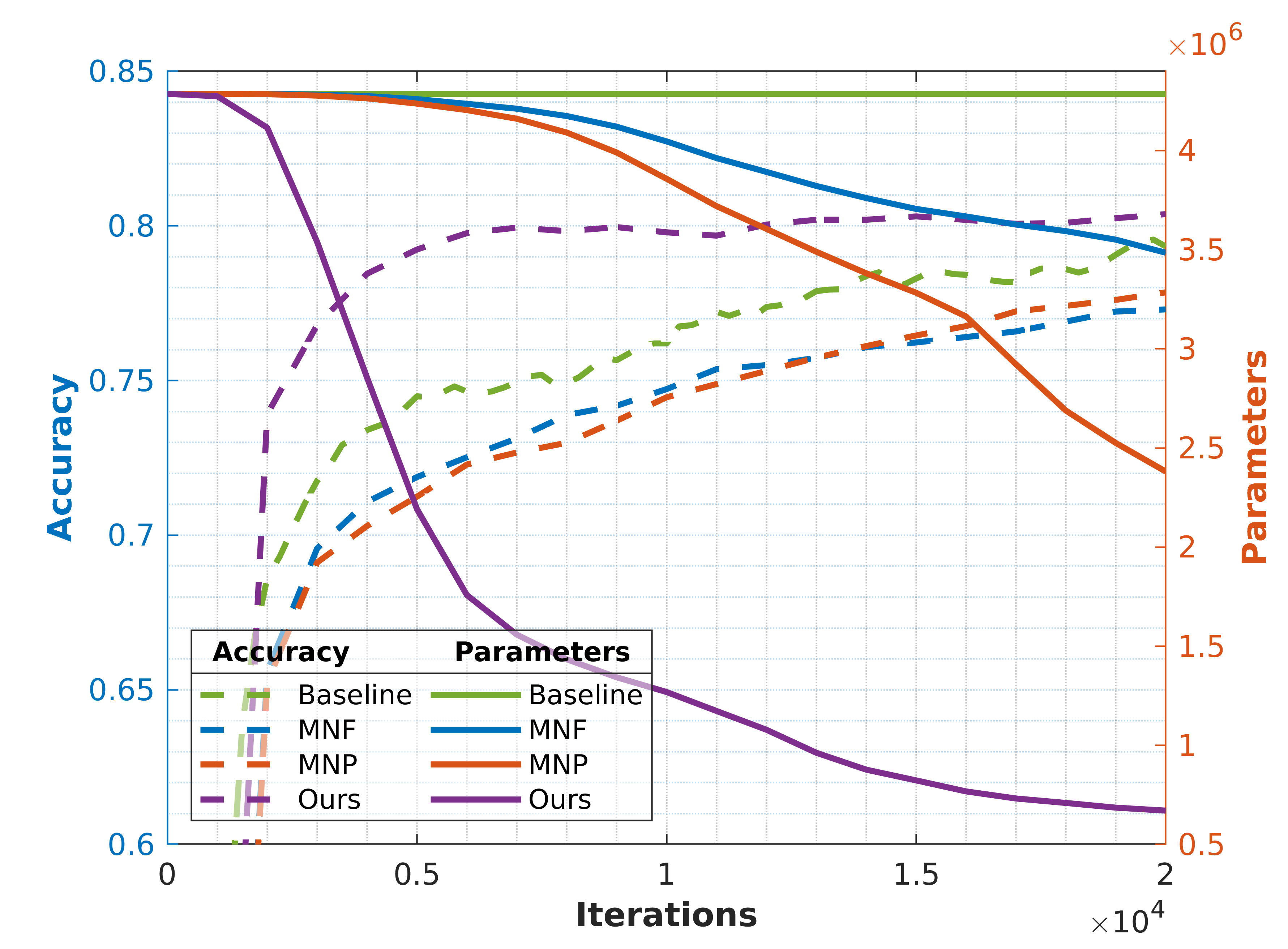}
\caption{Results on CIFAR-10}
\label{fig_CIFAR10}
\vspace*{0.5em}
\end{subfigure}
\begin{subfigure}{\textwidth}
\captionsetup{font=normalsize}
\centering
\includegraphics[width=0.45\linewidth,height=5.0cm]{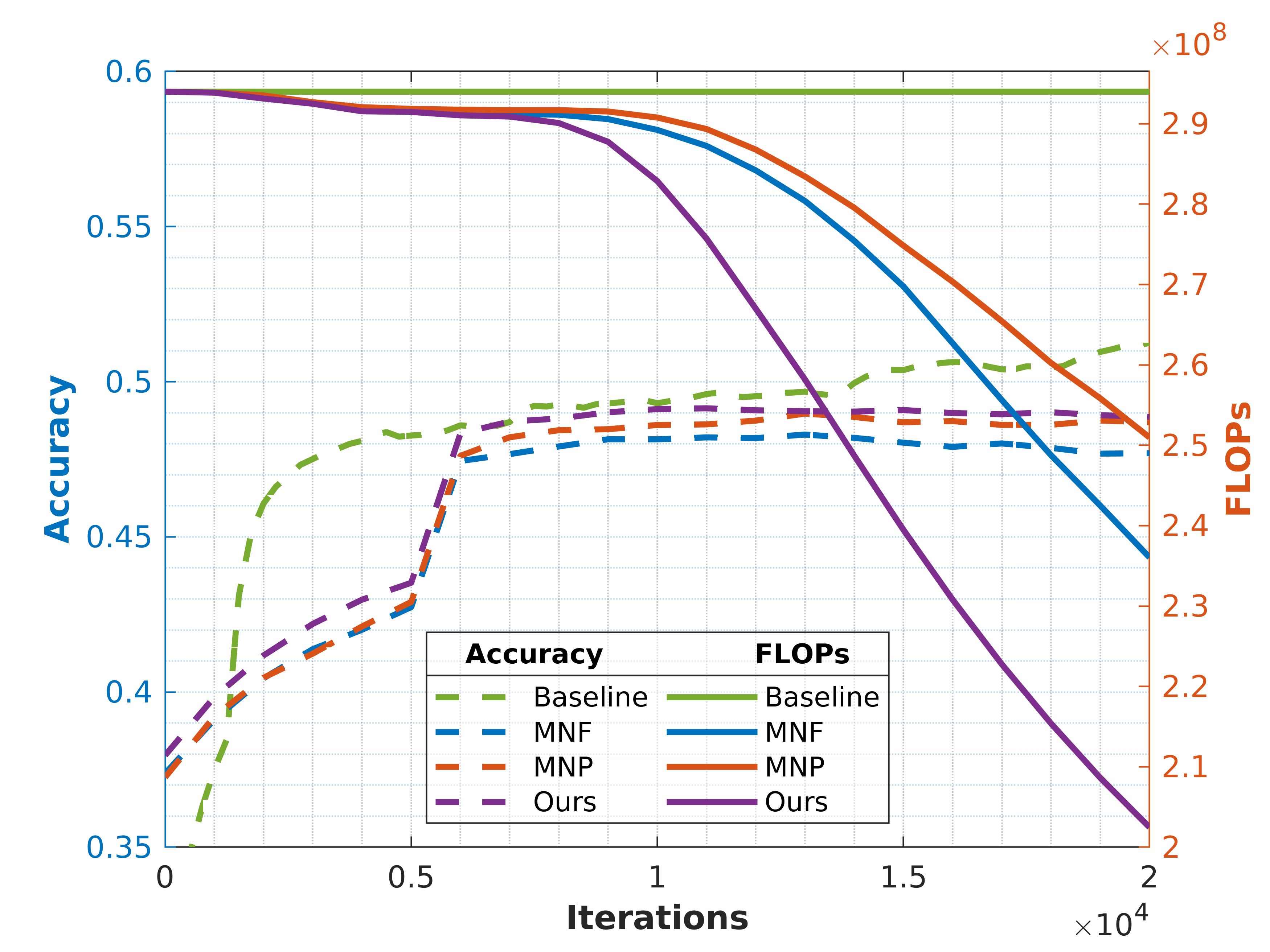}
\includegraphics[width=0.45\linewidth,height=5.0cm]{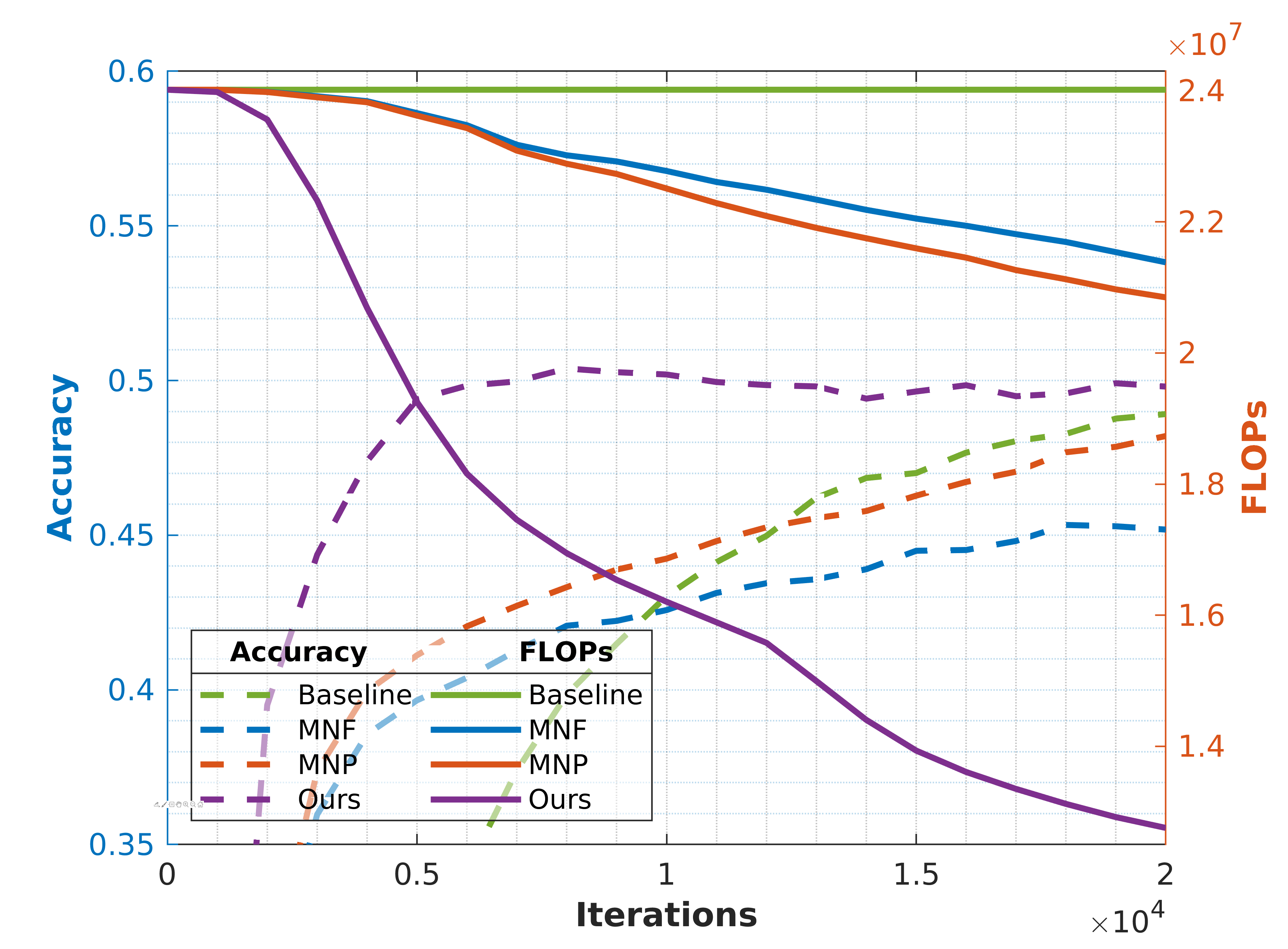}

\includegraphics[width=0.45\linewidth,height=5.0cm]{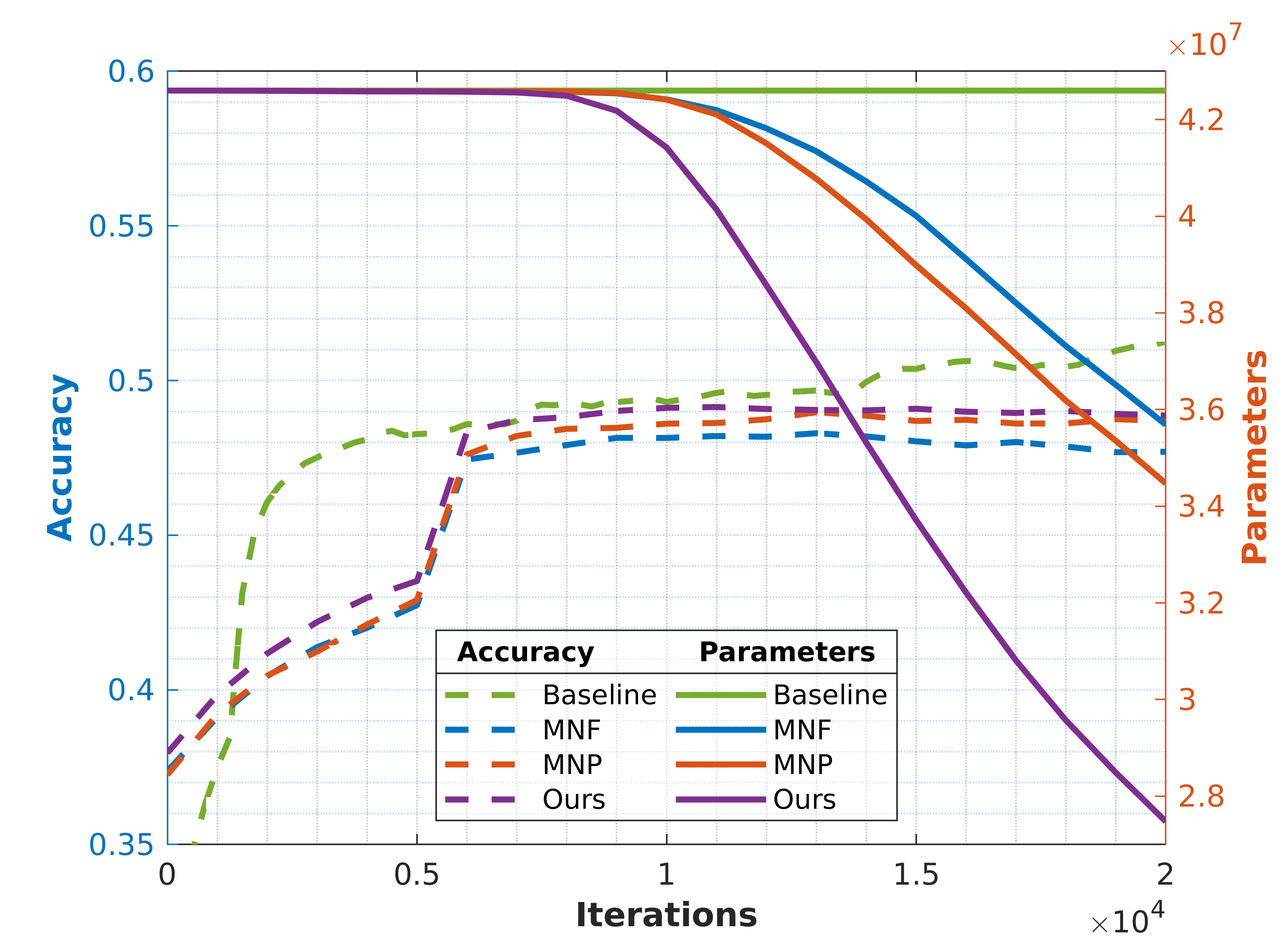}
\includegraphics[width=0.45\linewidth,height=5.0cm]{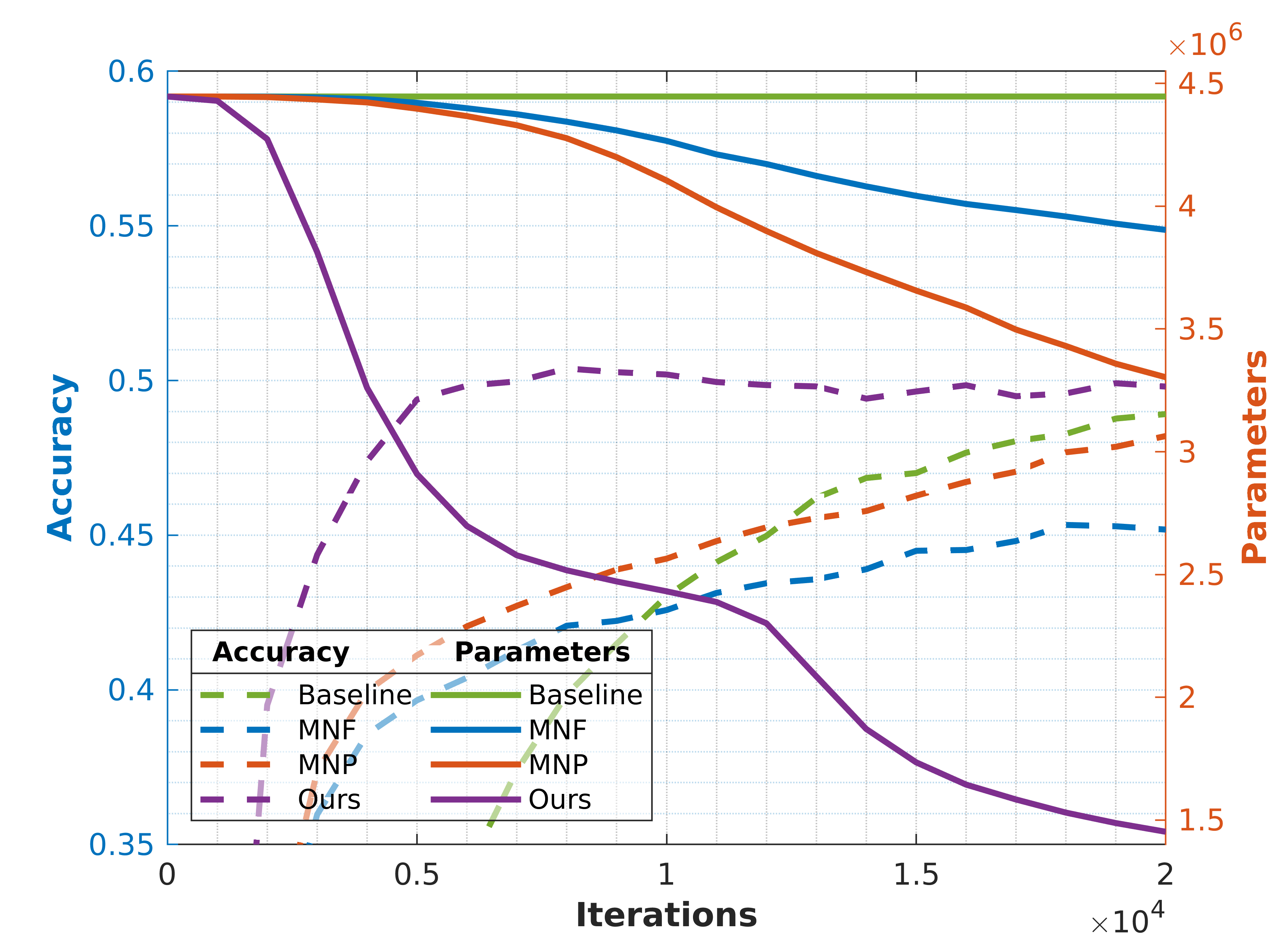}
\caption{Results on CIFAR-100}
\label{fig_CIFAR100}
\end{subfigure}
\caption{We compare FLOPs and model-parameters reduction trend for CIFAR-10 and CIFAR-100 benchmarks considering ResNet101 (left) and MobileNet\_v2 (right) as backbone networks. \textbf{MNF} and \textbf{MNP} are variants of existing method to exclusively optimize network structure  for FLOPs and model-parameters, respectively. For ResNet101, our method outperforms the existing method in FLOPs and model-parameters reduction with slightly better model performance. Especially, for already compact network MobileNet\_v2, our method brings superior network compression even with accuracy higher than the baseline.}
\label{fig_CIFAR}
\end{figure*}

\begin{table*}
\captionsetup{font=normalsize}
\vspace*{2.0em}
\caption{The results are reported on CIFAR-10 and CIFAR-100 with two different backbone CNNs. \boldmath$\alpha$=regularization-strength, \textbf{ACC}=accuracy on testset ($\%$), \textbf{RED}=reduction achieved ($\%$), \textbf{MNF},\textbf{MNP}=existing methods. Our proposed method offers an optimal solution for both FLOPs and model-parameters reductions, so it presents two RED columns, accordingly to the optimization considered. Our method outperforms existing one in terms of compression, with comparable or marginally lower accuracy (cases in \textbf{bold}) or even with higher accuracy (cases in \textbf{\underline{bold}}). Results show proposed framework's consistency, robustness, and generalizability.}
\begin{subtable}{\textwidth}
\centering
\captionsetup{font=normalsize}
\caption{\textbf{Results on MobileNet\_v2.} CIFAR-10 Baseline - ACC:84.9, FLOPs:\num{2.37e+7}, Model-Size:\num{4.29e+06}. CIFAR-100 Baseline - ACC:55.1, FLOPs:\num{2.40e+7}, Model-Size:\num{4.45e+06}.}
\begin{tabular}{ |c|c| c  c | c  c || c | c  c |}
\hline
& & \multicolumn{4}{ c|| }{\textbf{Optimization for FLOPs}} & \multicolumn{3}{ c| }{\textbf{Optimization for model-parameters}} \\
\hline
& \multirow{2}{*}{$\alpha$} & \multicolumn{2}{ c| }{MNF} & \multicolumn{3}{ c| }{\textbf{Ours}} & \multicolumn{2}{ c| }{MNP} \\
\cline{3-9}
&  & \multicolumn{1}{ c }{ACC} & \multicolumn{1}{ c| }{RED} & \multicolumn{1}{ c }{ACC} & \multicolumn{1}{ c|| }{RED} & \multicolumn{1}{ c| }{RED} & \multicolumn{1}{ c }{ACC} & \multicolumn{1}{ c| }{RED} \\
\hline 
\multirow{6}{*}{\rotatebox[origin=c]{90}{\textbf{CIFAR-10}}}  
&    1   &   77.8$\pm$0.8   &   6.0$\pm$0.1    &   82.8$\pm$0.2   &   \textbf{\underline{23.9}}$\pm$0.2  &  \textbf{\underline{55.6}}$\pm$0.4   &   77.6$\pm$0.1   &   9.4$\pm$0.3 \\ 
&    5   &   77.8$\pm$0.9   &   7.6$\pm$0.2    &   82.1$\pm$0.3   &   \textbf{\underline{39.6}}$\pm$0.2  &  \textbf{\underline{73.0}}$\pm$0.2   &   77.6$\pm$0.5   &   16.0$\pm$0.6 \\ 
&   10   &   77.2$\pm$0.4   &   9.5$\pm$0.2    &   81.4$\pm$0.2   &   \textbf{\underline{48.9}}$\pm$0.9  &  \textbf{\underline{78.8}}$\pm$0.7   &   78.0$\pm$0.8   &   31.0$\pm$1.9 \\ 
&   15   &   77.5$\pm$0.4   &   11.8$\pm$0.3   &   80.2$\pm$0.4   &   \textbf{\underline{53.1}}$\pm$1.6  &  \textbf{\underline{81.4}}$\pm$1.2   &   78.2$\pm$0.2   &   39.9$\pm$0.3 \\ 
&   20   &   77.4$\pm$0.1   &   14.7$\pm$0.1   &   80.3$\pm$0.2   &   \textbf{\underline{58.3}}$\pm$0.5  &  \textbf{\underline{84.3}}$\pm$0.1   &   77.8$\pm$0.5   &   44.4$\pm$0.6 \\ 
&   25   &   76.9$\pm$0.3   &   17.3$\pm$0.3   &   79.5$\pm$1.2   &   \textbf{\underline{61.5}}$\pm$0.2  &  \textbf{\underline{86.2}}$\pm$0.1   &   77.9$\pm$0.4   &   48.2$\pm$0.4 \\ 
\hline
\multirow{6}{*}{\rotatebox[origin=c]{90}{\textbf{CIFAR-100}}}
&    1   &   46.8$\pm$0.4   &   5.4$\pm$0.0    &   55.6$\pm$0.2   &   \textbf{\underline{10.5}}$\pm$0.3  &    \textbf{\underline{18.7}}$\pm$0.9   &   47.2$\pm$0.7   &   6.4$\pm$0.4 \\ 
&    5   &   46.5$\pm$0.5   &   6.1$\pm$0.1    &   53.6$\pm$0.4   &   \textbf{\underline{20.8}}$\pm$0.4  &    \textbf{\underline{36.0}}$\pm$0.4   &   47.3$\pm$0.9   &   10.0$\pm$0.6 \\ 
&   10   &   46.7$\pm$0.4   &   6.9$\pm$0.1    &   52.1$\pm$0.4   &   \textbf{\underline{29.1}}$\pm$0.3  &    \textbf{\underline{48.8}}$\pm$0.3   &   47.4$\pm$0.2   &   14.6$\pm$0.2 \\ 
&   15   &   46.5$\pm$0.6   &   7.8$\pm$0.5    &   51.0$\pm$0.5   &   \textbf{\underline{35.4}}$\pm$0.7  &    \textbf{\underline{56.5}}$\pm$0.9   &   47.1$\pm$0.5   &   18.3$\pm$1.3 \\ 
&   20   &   45.1$\pm$0.8   &   9.1$\pm$0.3    &   50.5$\pm$0.7   &   \textbf{\underline{39.9}}$\pm$1.1  &    \textbf{\underline{61.4}}$\pm$0.9   &   48.0$\pm$1.0   &   21.7$\pm$1.6 \\ 
&   25   &   45.3$\pm$1.1   &   11.0$\pm$0.6   &   49.6$\pm$0.6   &   \textbf{\underline{46.9}}$\pm$0.3  &    \textbf{\underline{67.3}}$\pm$0.5   &   48.6$\pm$0.0   &   25.7$\pm$0.8 \\ 
\hline
\end{tabular}
\vspace*{1.5em}
\label{res_MobileNetV2}
\end{subtable}

\begin{subtable}{\textwidth}
\centering
\captionsetup{font=normalsize}
\caption{\textbf{Results on ResNet101.} CIFAR-10 Baseline - ACC:80.3, FLOPs:\num{2.94e+8}, model-parameters:\num{4.24e+07}. CIFAR-100 Baseline - ACC:52.2, FLOPs:\num{2.94e+8}, model-parameters:\num{4.26e+07}.}
\begin{tabular}{ |c|c| c  c | c  c || c | c  c |}
\hline
& & \multicolumn{4}{ c|| }{\textbf{Optimization for FLOPs}} & \multicolumn{3}{ c| }{\textbf{Optimization for model-parameters}} \\
\hline
& \multirow{2}{*}{$\alpha$} & \multicolumn{2}{ c| }{MNF} & \multicolumn{3}{ c| }{\textbf{Ours}} & \multicolumn{2}{ c| }{MNP} \\
\cline{3-9}
&  & \multicolumn{1}{ c }{ACC} & \multicolumn{1}{ c| }{RED} & \multicolumn{1}{ c }{ACC} & \multicolumn{1}{ c|| }{RED} & \multicolumn{1}{ c| }{RED} & \multicolumn{1}{ c }{ACC} & \multicolumn{1}{ c| }{RED} \\
\hline 
\multirow{6}{*}{\rotatebox[origin=c]{90}{\textbf{CIFAR-10}}} 
& 1  &   79.9$\pm$0.6  &   2.0$\pm$0.4   &   79.8$\pm$2.0   &   \textbf{4.4}$\pm$0.5   &   \textbf{7.4}$\pm$0.7    &   80.2$\pm$0.4   &   5.9$\pm$0.7 \\
& 5  &   78.7$\pm$0.6  &   14.1$\pm$3.8  &   77.3$\pm$0.6   &   \textbf{18.3}$\pm$0.4  &   \textbf{27.9}$\pm$0.6   &   79.8$\pm$0.9   &   26.7$\pm$1.2 \\
& 10 &   77.4$\pm$0.9  &   27.1$\pm$4.9  &   78.6$\pm$1.3   &   \textbf{\underline{34.8}}$\pm$4.8  &   \textbf{\underline{45.8}}$\pm$4.5   &   77.2$\pm$1.7   &   31.0$\pm$6.7 \\
& 15 &   76.7$\pm$0.5  &   35.4$\pm$0.8  &   77.1$\pm$0.5   &   \textbf{\underline{41.8}}$\pm$2.1  &   \textbf{53.9}$\pm$1.8   &   77.6$\pm$2.5   &   48.1$\pm$6.8 \\
& 20 &   76.0$\pm$3.1  &   43.5$\pm$7.5  &   77.5$\pm$0.7   &   \textbf{50.0}$\pm$6.7  &   \textbf{61.6}$\pm$5.3   &   78.6$\pm$1.3   &   56.1$\pm$2.5 \\
& 25 &   76.3$\pm$0.9  &   48.7$\pm$1.6  &   76.3$\pm$1.6   &   \textbf{\underline{55.3}}$\pm$2.6  &   \textbf{\underline{66.9}}$\pm$1.7   &   76.0$\pm$2.2   &   58.5$\pm$1.3 \\[0.2ex] 
\hline
\multirow{6}{*}{\rotatebox[origin=c]{90}{\textbf{CIFAR-100}}}
& 1  &   50.5$\pm$2.4  &   1.0$\pm$0.1   &   50.3$\pm$0.8   &   1.0$\pm$0.0            &   0.1$\pm$0.0             &   50.0$\pm$0.9   &   0.1$\pm$0.0 \\
& 5  &   50.8$\pm$2.0  &   4.2$\pm$1.7   &   50.7$\pm$1.1   &   \textbf{7.8}$\pm$3.2  &   \textbf{\underline{8.1}}$\pm$3.6    &   49.9$\pm$0.5   &   2.0$\pm$0.4 \\
& 10 &   48.6$\pm$1.0  &   8.1$\pm$1.3   &   48.4$\pm$0.4   &   \textbf{17.7}$\pm$2.0 &   \textbf{\underline{20.5}}$\pm$2.0   &   47.9$\pm$1.1   &   10.3$\pm$2.9 \\
& 15 &   46.7$\pm$0.4  &   19.7$\pm$4.6  &   48.1$\pm$0.4   &   \textbf{\underline{31.1}}$\pm$3.5 &   \textbf{35.5}$\pm$3.2   &   48.5$\pm$0.6   &   19.1$\pm$5.3 \\
& 20 &   47.2$\pm$1.2  &   22.2$\pm$1.7  &   46.2$\pm$1.0   &   \textbf{32.1}$\pm$1.4 &   \textbf{38.7}$\pm$1.3   &   48.0$\pm$1.7   &   26.0$\pm$4.6 \\
& 25 &   44.5$\pm$2.3  &   29.7$\pm$2.8  &   47.5$\pm$1.3   &   \textbf{\underline{43.8}}$\pm$3.4 &   \textbf{\underline{51.1}}$\pm$3.1   &   45.9$\pm$2.2   &   34.9$\pm$1.4 \\
\hline
\end{tabular}
\label{res_ResNet101}
\end{subtable}
\label{res_all}
\end{table*}

\section{Experiments}\label{exps} 
We choose two widely used networks namely ResNet101 \cite{he2016deep} and MobileNet\_v2 \cite{sandler2018mobilenetv2} to examine the generalizability of proposed method on varied architecture designs. These networks were hand-crafted to achieve two distinct goals: the former was designed to obtain high accuracy while the latter was designed to yield low computation expense on mobile devices. Thus, we perform an extensive evaluation with these networks to show the effectiveness and adaptability of the proposed method (see Table~\ref{res_all}). We evaluate our proposed method on three standard classification datasets: CIFAR-10 \cite{krizhevsky2009learning}, CIFAR-100 \cite{krizhevsky2009learning} and Imagenet \cite{deng2009imagenet}. CIFAR-10 and CIFAR-100 datasets consist of $60000$ tiny images of dimensions 32$\times$32, categorized into $10$ and $100$ distinct classes respectively. Both datasets comprise $50K$ images in the trainset and $10K$ images in the testset. The ImageNet dataset consists of $1M$ images in train and $50K$ images in test split of dimensions 224$\times$224, categorized into $1000$ classes. In experiments, the teacher network is trained on target datasets in advance and kept fixed throughout the training process. Subsequently, an identical network architecture is considered as a student to discover its compact structure form capable of delivering competitive performance. For resource-aware optimization, we empirically divide the end-to-end optimization task into two sub-tasks by partitioning the network weights into two groups -- lower half and upper half, although our method is generalizable to a higher number of sub-tasks. 

Our implementation is based on TensorFlow and open-source tool MorphNet \cite{gordon2018morphnet}. We use Adam optimizer with a fixed learning rate of $10^{-4}$ for CIFAR-10 and CIFAR-100 and RMSProp optimizer for ImageNet with an initial learning rate of $10^{-4}$ that decays by a factor of 0.98 every 2.5 epochs. In each experiments, the student network is trained for $20K$ iterations with a mini-batch size of $100$ for CIFAR-10 and CIFAR-100. For ImageNet, ResNet101 and  MobileNet\_v2 models are trained for $500K$ iterations with mini-batch size of $32$. In the network structure learning process, fixed number of iterations is essential for comparison - \textit{i.e.} more or less iterations can lead to different network structure and performance. We used different values of regularization-strength $\alpha$ with fixed $\gamma=0.01$ to guarantee fair comparison with \cite{gordon2018morphnet}. For distilling knowledge, we use $T=10$ and $\lambda=0.5$. After training, the performance of the student network is evaluated over the entire test-set to observe the effectiveness of learned lightweight network structure. We compare our proposed method with the stand-alone variants of MorphNet, namely MNF (entire network optimized for FLOPs) and MNP (entire network optimized for model-parameters). The results are reported in terms of Accuracy (ACC) and the FLOPs/model-parameters Reduction (RED) metrics - \textit{i.e.} the percentage reduction with respect to the \textit{FLOPs} or \textit{model-parameters} cost of the teacher network. All parameters were kept consistent for entire experiments and results are averaged over three runs.

\begin{figure*}
\centering
\includegraphics[width=0.45\linewidth,height=5cm]{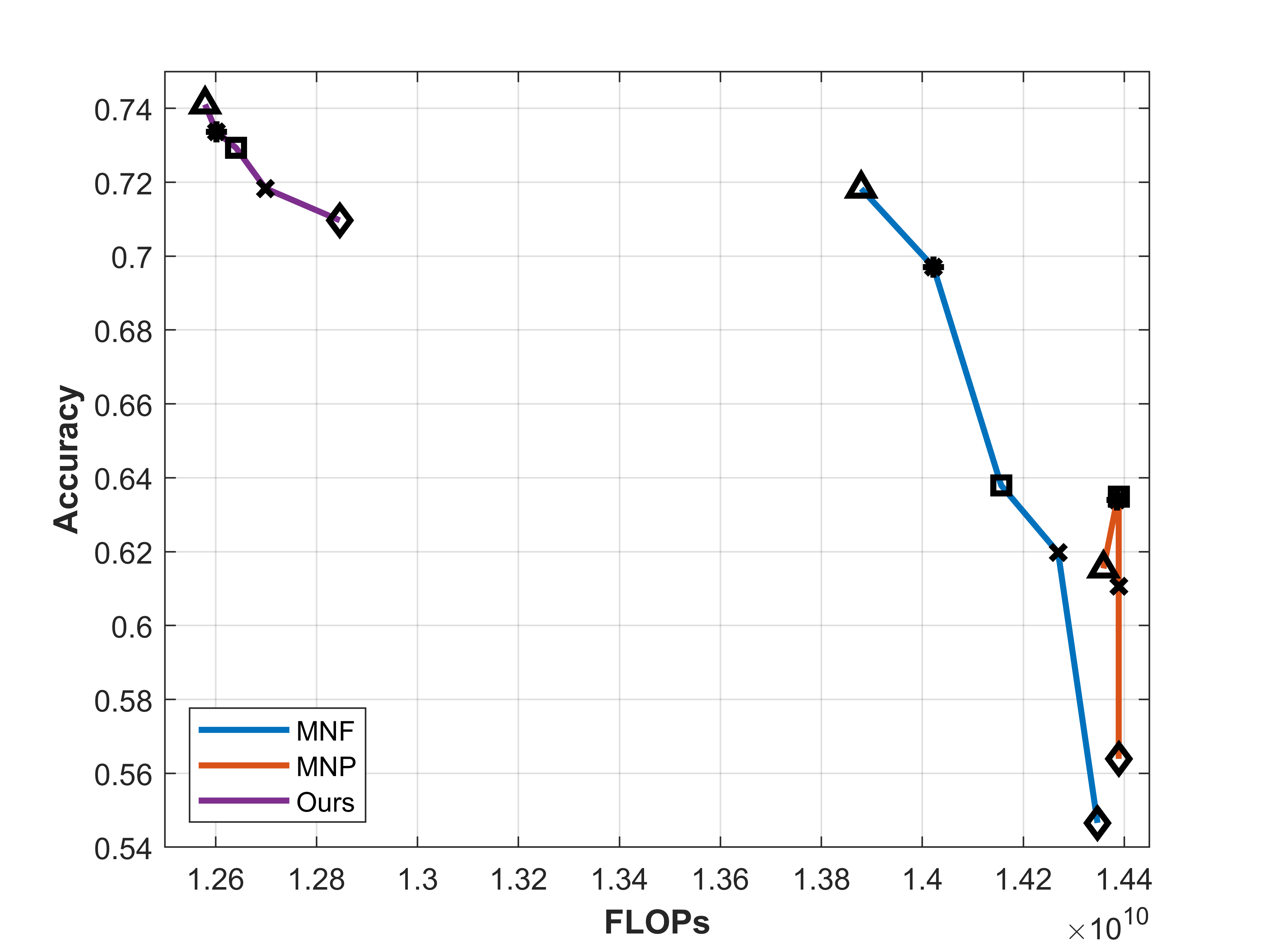}
\includegraphics[width=0.45\linewidth,height=5cm]{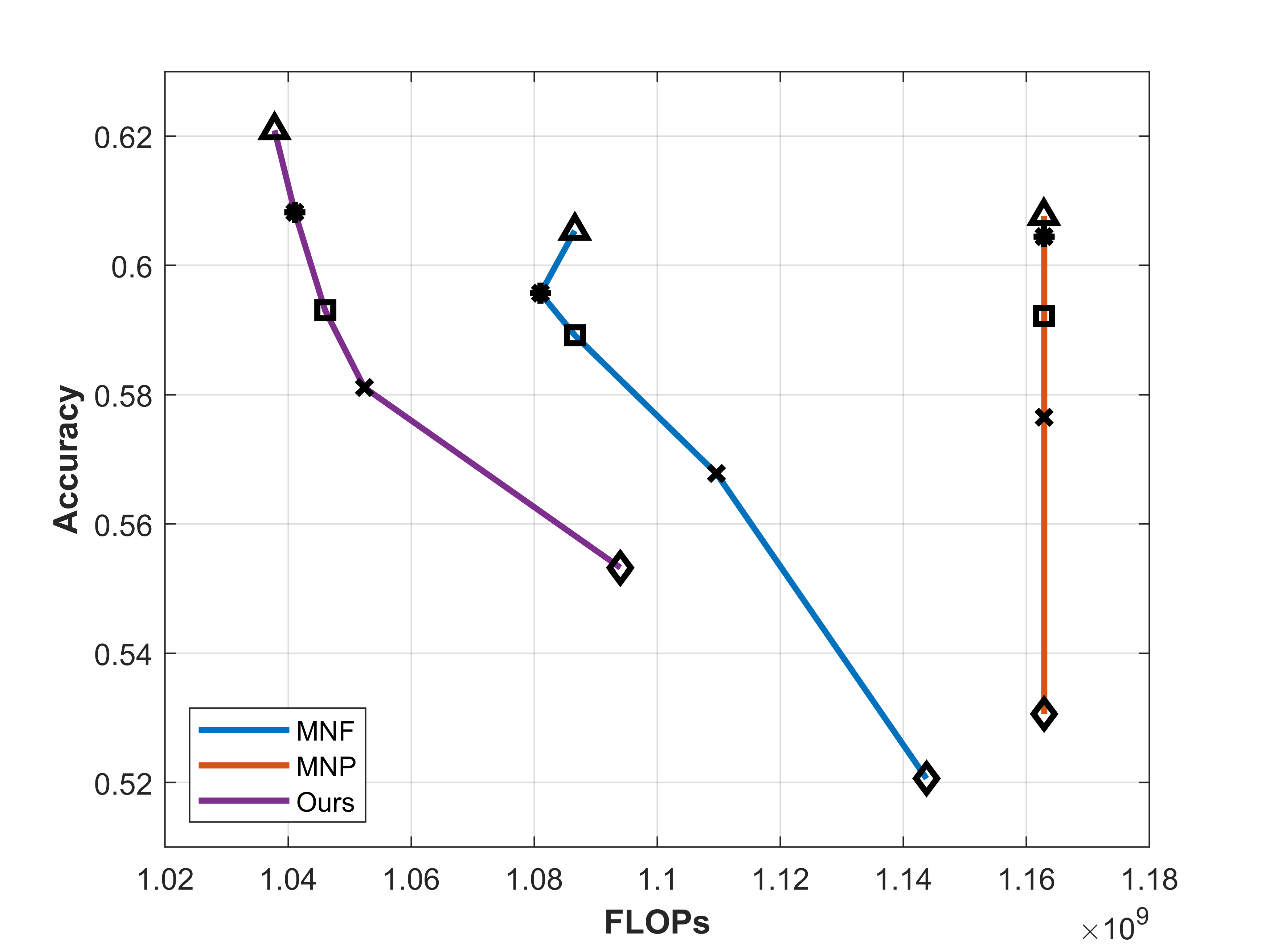}

\includegraphics[width=0.45\linewidth,height=5cm]{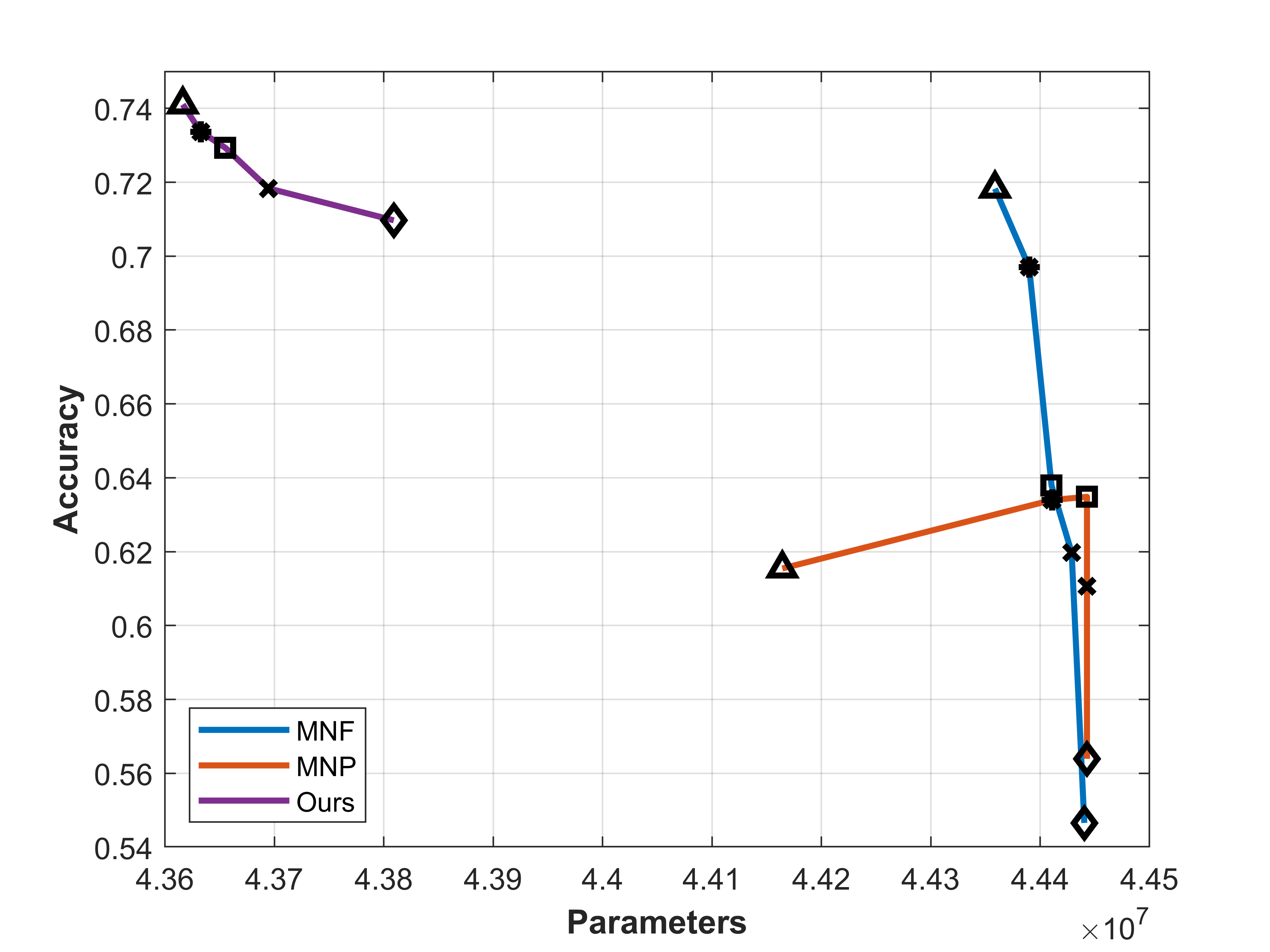}
\includegraphics[width=0.45\linewidth,height=5cm]{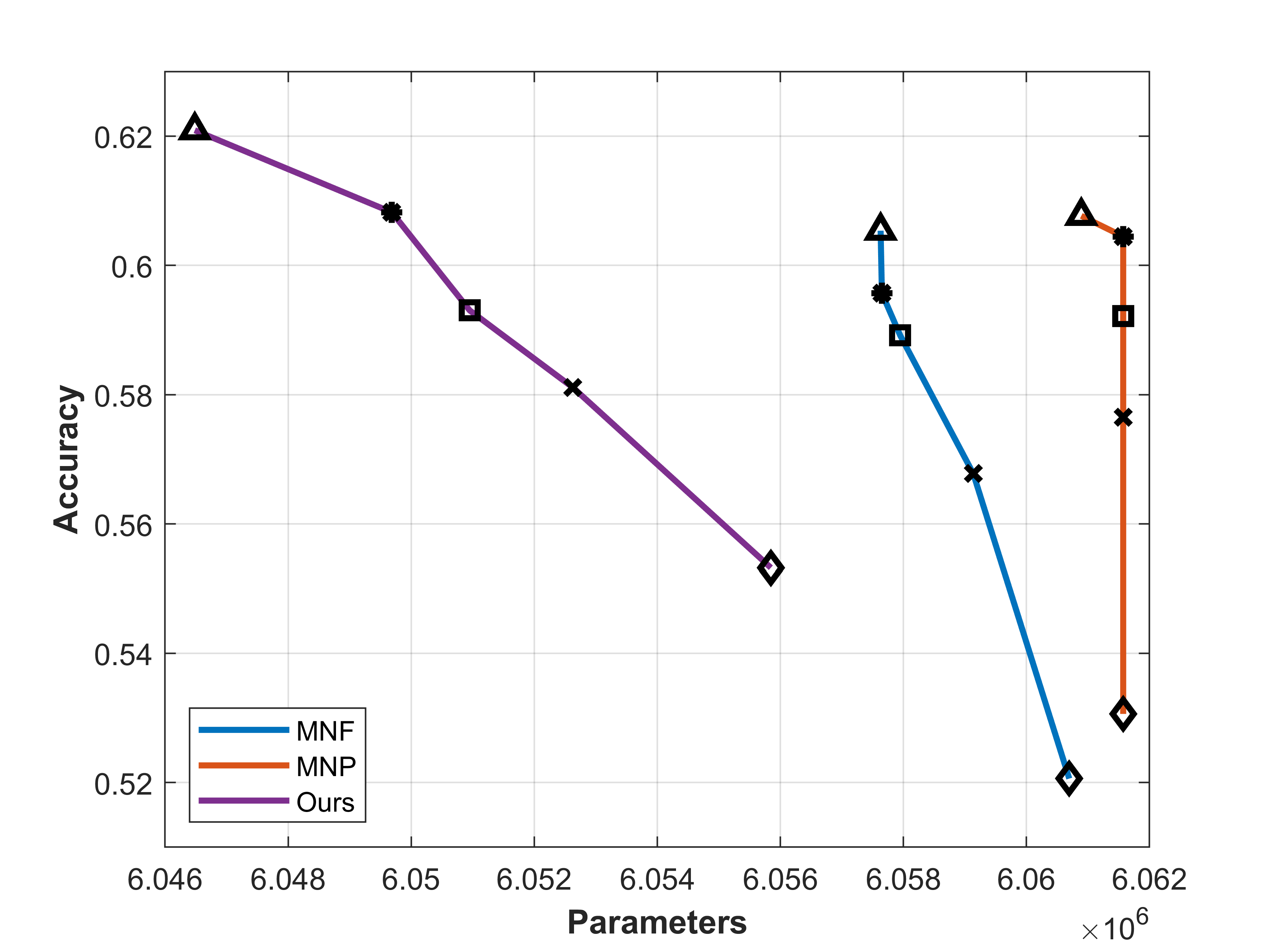}
\caption{\textbf{Results on ImageNet.} We compare FLOPs and model-parameters reduction trend for ResNet101 (left) and MobileNet\_v2 (right). \textbf{MNF} and \textbf{MNP} are variants of existing method to optimize network structure for FLOPs and model-parameters, respectively. We present accuracy vs. FLOPs/parameters results at $100, 200, 300, 400$ and $500$ thousand iterations marked as $\diamond, \times, \square, *$ and $\triangle$, respectively. For ResNet101, our method outperforms the existing method by a large margin in terms of FLOPs and model-parameters reduction with superior model performance.}
\label{fig_ImageNet}
\end{figure*}
%
%

\section{Results}\label{result_sec}
In this Section we demonstrate that our method consistently outperforms the existing method by substantial margins, both, in terms of FLOPs and model-parameters reduction while offering better model performance. Targeting networks of different capacities, we present results for the two image recognition datasets CIFAR-10 and CIFAR-100 in Table~\ref{res_all} with varying compression intensity steered by the parameter $\alpha$. In each sub-caption, we report the performance achieved by the teacher network in terms of accuracy and original \textit{FLOPs}/\textit{model-parameters} cost after been trained for the same image recognition task. We compare the proposed method with the stand-alone MorphNet MNF (entire network optimized for FLOPs) and MNP (entire network optimized for model-parameters). We demonstrate that our method brings better compression-performance tradeoff over all regularization strength $\alpha$ considered. 

In particular, evolution trend during training for CIFAR-10 in Figure~\ref{fig_CIFAR10} (right) shows that our method is relatively more effective for an already compact network (MobileNet\_v2) with $2\times$ better FLOPs reduction than MNF and $5.2\times$ better model-parameters reduction than MNP.
Such substantial gain in compression is achieved with up to $1.05\%$ better recognition accuracy than the baseline teacher network. The most notable difference in trends is that our method brings superior model compression and performance right from the beginning and keeps on fine-tuning over successive iterations.  
Similarly, for ResNet101 in Figure~\ref{fig_CIFAR10} (left), our method learns network structure capable of delivering slightly better performance while being $1.1\times$ and $1.3\times$ more compressed in terms of FLOPs and model-parameters, respectively, than the existing method.

Also, Figure~\ref{fig_CIFAR100} shows the same consistent trend for CIFAR-100 on both networks. For MobileNet\_v2 in Figure~\ref{fig_CIFAR100} (right) our method brings $1.7\times$ and $2.7\times$ better compression in terms of FLOPs and model-parameters, respectively with $0.88\%$ better recognition accuracy than the baseline.
For ResNet101 in Figure~\ref{fig_CIFAR100} (left), our method learns network structure capable of delivering slightly better performance while being $1.2\times$ and $1.3\times$ more compressed in terms of FLOPs and model-parameters, respectively, in comparison with the existing method.

Finally, we report experiments considering the popular large scale ImageNet dataset on both networks. Figure~\ref{fig_ImageNet} shows the accuracy versus FLOPs (top row) and model-parameters (bottom row) reduction. We present results achieved after $500K$ iterations with $\triangle$ along with intermediate results (\textit{i.e.} results after $100, 200, 300$ and $400$ thousand iterations marked as $\diamond, \times, \square$ and $*$, respectively) obtained during network structure learning. As expected from the insights discussed in Section~\ref{block}, the proposed method works best in terms of learning optimum network structure. Since lower layers are optimized for FLOPs and higher layers for model-parameters, the structure of each block is optimized accordingly to the most suitable resource constraint. For MobileNet\_v2 in Figure~\ref{fig_ImageNet} (right), our method brings outstanding model compression along with higher classification accuracy even in the cases where MNP has not started the optimization yet. A similar trend is also confirmed for ResNet101 in Figure~\ref{fig_ImageNet} (left) in which our method obtains superior model compression after only $100K$ iterations that is way higher than what is achieved after $500K$ iterations using the existing method.

\section{Conclusion}\label{conclus}
In this paper, we present a resource-aware network structure learning method, which enables suitable optimization in different sections of the seed network considering FLOPs and model-parameters constraints - \textit{i.e.} lower layers are optimized for FLOPs and higher layers for model-parameters. Furthermore, Our method leverages privileged information to impose control over predictions to preserve high-quality model performance. In an extensive evaluation of various network architectures and datasets, our method brings state of the art network compression that outperforms the existing method by a large margin while maintaining better control over the compression-performance tradeoff.






\bibliographystyle{IEEEtran}
\bibliography{egbib}
%
%

\end{document}